\documentclass{article}

\usepackage{arxiv}

\usepackage[utf8]{inputenc} % allow utf-8 input
\usepackage[T1]{fontenc}    % use 8-bit T1 fonts
\usepackage{hyperref}       % hyperlinks
\usepackage{url}            % simple URL typesetting
\usepackage{booktabs}       % professional-quality tables
\usepackage{amsfonts}       % blackboard math symbols
\usepackage{nicefrac}       % compact symbols for 1/2, etc.
\usepackage{microtype}      % microtypography
\usepackage{graphicx}
\usepackage[square,numbers]{natbib}
\usepackage{doi}
\usepackage{multirow}

\title{Explainable AI (XAI): A Systematic Meta-Survey of Current Challenges and Future Opportunities}

\author{ {\hspace{1mm}Waddah Saeed} \\
	Center for Artificial Intelligence Research\\
	University of Agder\\
	Grimstad, Norway \\
	\texttt{waddah.waheeb@uia.no} \\
	\And
	{\hspace{1mm}Christian Omlin} \\
	Center for Artificial Intelligence Research\\
	University of Agder\\
	Grimstad, Norway \\
	\texttt{christian.omlin@uia.no} \\
}

\renewcommand{\shorttitle}{Explainable AI: A Systematic Meta-Survey of Current Challenges and Future Opportunities}

\hypersetup{
pdftitle={Explainable AI (XAI): A Systematic Meta-Survey of Current Challenges and Future Opportunities},
pdfauthor={Waddah ~Saeed, Christian ~Omlin},
pdfkeywords={Explainable AI (XAI), interpretable, black-box, machine learning, deep learning, challenges, research directions, meta-survey}}

\begin{document}
\maketitle

\begin{abstract}
The past decade has seen significant progress in artificial intelligence (AI), which has resulted in algorithms being adopted for resolving a variety of problems. However, this success has been met by increasing model complexity and employing black-box AI models that lack transparency. In response to this need, Explainable AI (XAI) has been proposed to make AI more transparent and thus advance the adoption of AI in critical domains. Although there are several reviews of XAI topics in the literature that identified challenges and potential research directions in XAI, these challenges and research directions are scattered. This study, hence, presents a systematic meta-survey for challenges and future research directions in XAI organized in two themes: (1) general challenges and research directions in XAI and (2) challenges and research directions in XAI based on machine learning life cycle's phases: design, development, and deployment. We believe that our meta-survey contributes to XAI literature by providing a guide for future exploration in the XAI area.

\end{abstract}

% keywords can be removed
\keywords{Explainable AI (XAI) \and interpretable \and black-box \and machine learning \and deep learning \and challenges \and research directions \and meta-survey}

\section{Introduction}
Artificial intelligence (AI) has undergone significant and continuous progress in the past decade, resulting in the increased adoption of its algorithms (e.g., machine learning (ML) algorithms) for solving many problems, even those that were difficult to resolve in the past. However, these outstanding achievements are accompanied by increasing model complexity and utilizing black-box AI models that lack transparency. Therefore, it becomes necessary to come up with solutions that can contribute to addressing such a challenge, which could help expand utilizing AI systems in critical and sensitive domains (e.g., healthcare and security domains) where other criteria must be met besides the high accuracy.

Explainable artificial intelligence (XAI) has been proposed as a solution that can help to move towards more transparent AI and thus avoid limiting the adoption of AI in critical domains \cite{Adadi2018,arrieta2020}. Generally speaking, according to \cite{gunning2016broad}, XAI focuses on developing explainable techniques that empower end-users in comperhending, trusting, and efficiently managing the new age of AI systems. Historically, the need for explanations dates back to the early works in explaining expert systems and Bayesian networks \cite{biran2017explanation}. Deep learning (DL), however, has made XAI a thriving research area.

Every year, a large number of studies dealing with XAI are published. At the same time, various review studies are published covering a range of general or specific aspects of XAI. With many of these review studies, several challenges and research directions are discussed. While this has led to identifying challenges and potential research directions, however, they are scattered.

To the best of our knowledge, this is the first meta-survey that explicitly organizes and reports on the challenges and potential research directions in XAI. This meta-survey aims to provide a reference point for researchers interested in working on challenges and potential research directions in XAI.

The organization of the paper is as follows. In Section 2, we discuss the need for XAI from various perspectives. Following that, Section 3 tries to contribute to a better distinction between explainability and interpretability. The protocol used in planning and executing this systematic meta-survey is presented in Section 4. Afterward, Section 5 discusses the challenges and research directions in XAI. Lastly, final remarks are highlighted in Section 6.

\section{Why Explainable AI is Needed?}
\label{motivations}

Nowadays, we are surrounded by black-box AI systems utilized to make decisions for us, as in autonomous vehicles, social networks, and medical systems. Most of these decisions are taken without knowing the reasons behind these decisions.

According to \cite{Adadi2018}, not all black-box AI systems need to explain why they take each decision because this could result in many consequences such as reducing systems efficiency and increasing development cost. Generally, explainability/interpretability is not needed in two situations \cite{Doshi-Velez2017-yb}: (1) results that are unacceptable are not accompanied by severe consequences, (2) the problem has been studied in-depth and well-tested in practice, so the decision made by the black-box system is trustworthy, e.g., advertisement system and postal code sorting. Therefore, we should think about why and when explanations/interpretations can be helpful \cite{Adadi2018}.

Based on the retrieved surveys in this work, the need for XAI can be discussed from various perspectives as shown in Fig. \ref{perspectives}. The perspective groups below are to some extent based on the work in \cite{Gade2020-gt}:

\begin{figure}[t]
  \centering
  \includegraphics[width=0.9\linewidth]{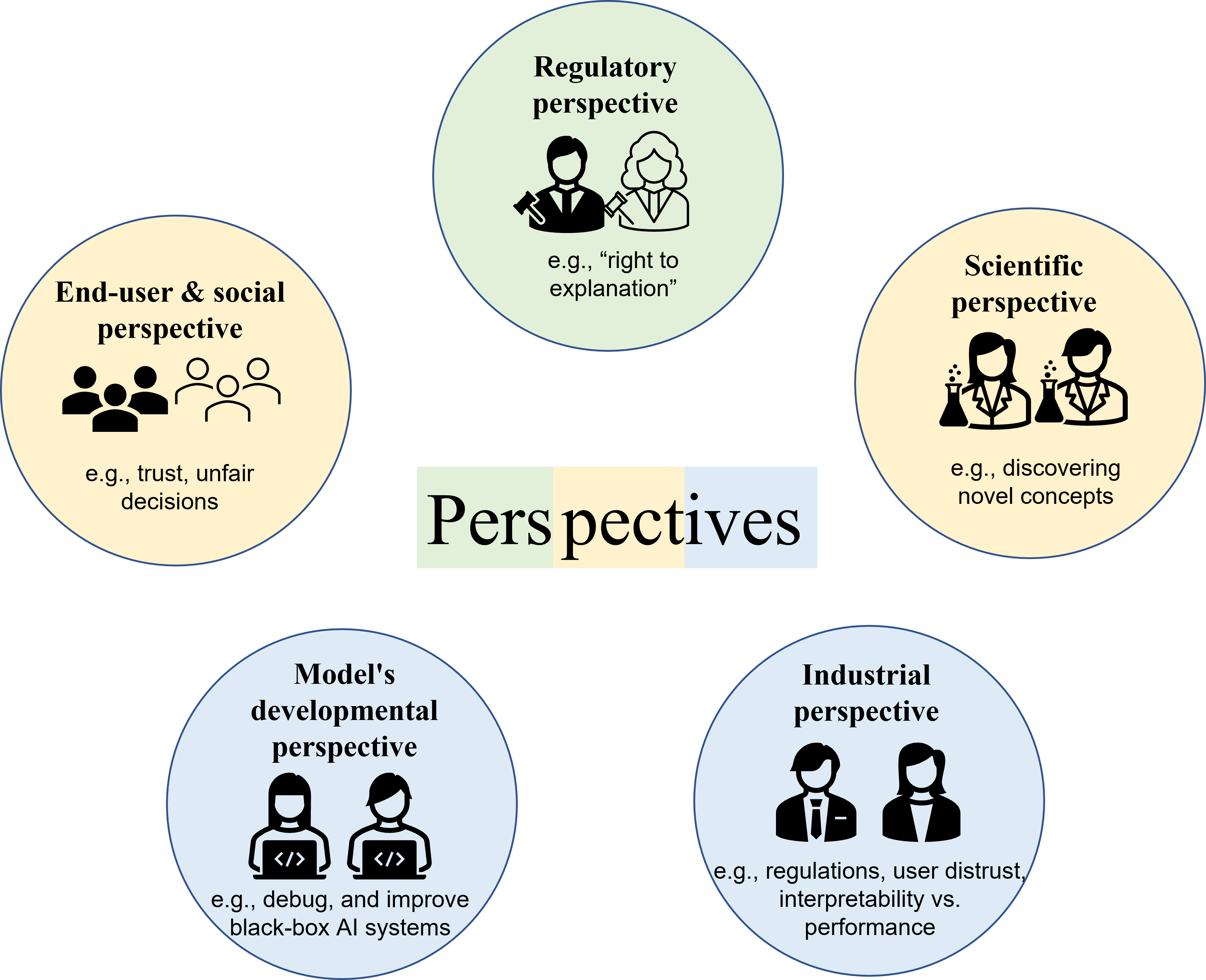}
  \caption{The five main perspectives for the need for XAI.}
  \label{perspectives}
\end{figure}

\begin{itemize}
\item Regulatory perspective: Black-box AI systems are being utilized in many areas of our daily lives, which could resulting in unacceptable decisions, especially those that may lead to legal effects. Thus, it poses a new challenge for the legislation. The European Union's General Data Protection Regulation (GDPR)\footnote{https://www.privacy-regulation.eu/en/r71.htm} is an example of why XAI is needed from a regulatory perspective. These regulations create what is called "right to explanation," by which a user is entitled to request an explanation about the decision made by the algorithm that considerably influences them \cite{GoodmanFlaxman2017}. For example, if an AI system rejects one’s application for a loan, the applicant is entitled to request justifications behind that decision to guarantee it is in agreement with other laws and regulations \cite{Samek2019-po}. However, the implementation of such regulations is not straightforward, challenging, and without an enabling technology that can provide explanations, the "right to explanation" is nothing more than a "dead letter" \cite{Samek2019-po,Payrovnaziri2020-xi,Guidotti2018-tx}.

\item Scientific perspective: When building black-box AI models, we aim to develop an approximate function to address the given problem. Therefore, after creating the black-box AI model, the created model represents the basis of knowledge, rather than the data \cite{Molnar2021-do}. Based on that, XAI can be helpful to reveal the scientific knowledge extracted by the black-box AI models, which could lead to discovering novel concepts in various branches of science.

\item Industrial perspective: Regulations and user distrust in black-box AI systems represent challenges to the industry in applying complex and accurate black-box AI systems \cite{Veiber2020-gv}. Less accurate models that are more interpretable may be preferred in the industry because of regulation reasons \cite{Veiber2020-gv}. A major advantage of XAI is that it can help in mitigating the common trade-off between model interpretability and performance \cite{arrieta2020}, thus meeting these common challenges. However, it can increase development and deployment costs.

\item Model's developmental perspective: Several reasons could contribute to inappropriate results for black-box AI systems, such as limited training data, biased training data, outliers, adversarial data, and model's overfitting. Therefore, what black-box AI systems have learned and why they make decisions need to be understood, primarily when they affect humans’ lives. For that, the aim will be to use XAI to understand, debug, and improve the black-box AI system to enhance its robustness, increase safety and user trust, minimize or prevent faulty behavior, bias, unfairness, and discrimination \cite{Confalonieri2021-yx}. Furthermore, when comparing models with similar performance, XAI can help in the selection by revealing the features that the models use to produce their decisions \cite{Samek2017-lb,arras2017relevant}. In addition, XAI can serve as a proxy function for the ultimate goal because the algorithm may be optimized for an incomplete objective \cite{Doshi-Velez2017-yb}. For instance, optimizing an AI system for cholesterol control with ignoring the likelihood of adherence \cite{Doshi-Velez2017-yb}.

\item End-user and social perspectives: In the literature of deep learning \cite{Guidotti2018-tx,szegedy2013intriguing,nguyen2015deep}, it has been shown that altering an image such that humans cannot observe the change can lead the model in producing a wrong class label. On the contrary, completely unrecognizable images of humans can be recognizable with high confidence using DL models. Such findings could raise doubts about trusting such black-box AI models \cite{Guidotti2018-tx}. The possibility to produce unfair decisions is another concern about black-box AI systems. This could happen in case black-box AI systems are developed using data that may exhibit human biases and prejudices \cite{Guidotti2018-tx}. Therefore, producing explanations and enhancing the interpretability of the black-box AI systems will help in increasing trust because it will be possible to understand the rationale behind the model’s decisions, and we can know if the system serves what it is designed for instead of what it was trained for \cite{Guidotti2018-tx,Burkart2021-tq}. Furthermore, the demand for the fairness of black-box AI systems' decisions, which cannot be ensured by error measures, often leads to the need for interpretable models \cite{Lipton2018-dx}.

\end{itemize}

The above list is far from complete, and there may be an overlap between these perspectives. However, these highlight the most critical reasons why XAI is needed.

\section{From Explainability to Interpretability}
\label{expvsintp}
In the literature, there seems to be no agreement upon what ``explainability'' or ``interpretability'' mean. While both terms are often used interchangeably in the literature, some examples from the selected papers distinguish them \cite{Markus2021-bp,arrieta2020,Akata2020-ct,Chakraborty,zhang2020survey,Chatzimparmpas}. To show that there are no agreement upon definitions, three different definitions from \cite{Markus2021-bp,arrieta2020,Akata2020-ct} are provided. In \cite{Markus2021-bp}, the authors stated that ``\emph{... we consider interpretability a property related to an explanation and explainability a broader concept referring to all actions to explain.}''. In another work \cite{arrieta2020}, interpretability is defined as ``\emph{the ability to explain or to provide the meaning in understandable terms to a human.}'', while ``\emph{explainability is associated with the notion of explanation as an interface between humans and a decision-maker that is, at the same time, both an accurate proxy of the decision-maker and comprehensible to humans...}''. Another distinction is drawn in \cite{Akata2020-ct}, in which authors stated that ``\emph{... In the case of interpretation, abstract concepts are translated into insights useful for domain knowledge (for example, identifying correlations between layers in a neural network for language analysis and linguistic knowledge). An explanation provides information that gives insights to users as to how a model came to a decision or interpretation.}''. It can be noticed from these distinctions that the authors have different definitions for these two terms. In addition, there is still considerable ambiguity in some of the given distinctions.

To contribute to a better distinction between explainability and interpretability, this paper attempts to present a distinction between these terms as follows:

\begin{quote}
Explainability provides \textbf{insights} to a \textbf{targeted audience} to fulfill a \textbf{need}, whereas interpretability is the degree to which the provided \textbf{insights} can make sense for the \textbf{targeted audience}'s domain knowledge.
\end{quote}

There are three components in the definition of explainability, as shown in the above distinction: \textbf{insights}, \textbf{targeted audience}, and \textbf{need}. \textbf{Insights} are the output from explainability techniques used (e.g., text explanation, feature relevance, local explanation). These insights are provided to a \textbf{targeted audience} such as domain experts (e.g., medical doctors), end-users (e.g., users affected by the model decision), modeling experts (e.g., data scientists). The \textbf{need} for the provided insights may be to handle any issues discussed in Section \ref{motivations} such as justifying decisions, discovering new knowledge, improving the black-box AI model, and ensuring fair decisions. That means explainability aims to help the targeted audience to fulfill a need based on the provided insights from the explainability techniques used.

As for interpretability, are the provided explanations consistent with the targeted audience’s knowledge? Do the explanations make sense to the targeted audience? Is the targeted audience able to reason/inference to support decision-making? Are the provided explanations reasonable for the model's decision?

Although the distinction is not ideal, we believe that it represents an initial step toward understanding the difference between explainability and interpretability. This paper will use this proposed distinction when discussing the challenges and research directions in XAI.

\section{Systematic review planning and execution}
\label{methodology}

This work is mainly based on a systematic literature review (SLR) introduced by Kitchenham and Charters \cite{keele2007}. We started our SLR by specifying the research question: \emph{What are the challenges and research directions in XAI reported in the existing survey studies?} The answer to this question will help researchers and practitioners to know the various dimensions that one can consider when working in the XAI research area. 

Having the research question established, the search terms based on the research question are: 

\begin{itemize}
\item XAI keywords: explainable, XAI, interpretable.
\item Review keywords: survey, review, overview, literature, bibliometric, challenge, prospect, agenda, trend, insight, opportunity, lesson, research direction
\end{itemize}

Relevant and important electronic databases were selected and used for searching the primary studies based on the search terms. These databases are:

\begin{enumerate}
\item Scopus
\item Web of Science
\item Science Direct
\item Institute of Electrical and Electronics Engineers Xplore Digital Library (IEEEXplore)
\item Springer Link
\item Association for Computing Machinery Digital Library (ACM)
\item Google Scholar
\item arXiv
\end{enumerate}

The last search using these databases was conducted on 16 Feb. 2021. After obtaining search results, all studies were analyzed individually to assess their relevance in the context of this SLR. Inclusion and exclusion criteria were used to select or discard the retrieved studies. The inclusion criteria are the following: 

\begin{itemize} 
\item The study presents a survey of explainable AI.
\item The study presents challenges and/or research directions for XAI.
\end{itemize}

On the other hand, the exclusion criteria are the following:

\begin{itemize} 
\item The study is not written in English. 
\item The study presents a survey of XAI without discussing any challenges or research directions.
\end{itemize}

The retrieved studies were first analyzed by their titles and abstracts to decide if the paper matched the first inclusion criterion. If matched, the paper was analyzed in detail in the second step. In the second step, the exclusion criteria and the second inclusion criterion were checked. 

We reviewed the list of references of the selected studies to include other papers that may not be retrieved from the selected electronic databases, which resulted in retrieving eight non-survey papers that reported challenges and/or research directions in XAI \cite{Aurangzeb,Doshi-Velez2017-yb,fox2017explainable,gunning2019xai,Lipton2018-dx,Molnar2021-do,Preece2018,Ras2018}.

Overall, the total number of selected papers is 58. The majority of the selected papers were published in 2020, as shown in Table \ref{year}. In 2021, we found 10 papers. However, since the last search was in Feb 2021, we expect more publications to appear until the end of the year. As shown in Table \ref{type}, the primary outlet for the selected papers are journal articles followed by conference papers and arXiv papers. The distribution of the selected papers per publisher is shown in Table \ref{publisher}.

\begin{table}[t]
\centering
\caption{Distribution of selected papers per year.\label{year}}
\begin{tabular}{lp{2.5cm}}
\toprule
Year & Number of papers\\
\midrule
2017 & 4\\
2018 & 12\\
2019 & 10\\
2020 & 22\\
2021 & 10\\
Total & 58\\
\bottomrule
\end{tabular}
\\
\end{table}

\begin{table}[t]
\centering
\caption{Distribution of selected papers per publication type.\label{type}}
\begin{tabular}{lp{2.5cm}}
\toprule
Publication type & Number of papers\\
\midrule
Journal articles & 28\\
Conference papers & 12\\
arXiv & 12\\
Book chapter & 5\\
Book & 1\\
Total & 58\\
\bottomrule
\end{tabular}
\\
\end{table}

\begin{table}[!t]
\centering
\caption{Distribution of selected papers per publisher.\label{publisher}}
\begin{tabular}{lp{2.5cm}}
\toprule
Publisher & Number of papers\\
\midrule
Springer & 12\\
arXiv & 12\\
IEEE & 9\\
Elsevier & 7\\
ACM & 6\\
Frontiers & 2\\
MDPI & 1\\
Wiley & 1\\
Sage & 1\\
Others & 7\\
Total & 58\\
\bottomrule
\end{tabular}
\\
\end{table}

\section{Discussion}
\begin{figure}
\centering
\includegraphics[width=\linewidth]{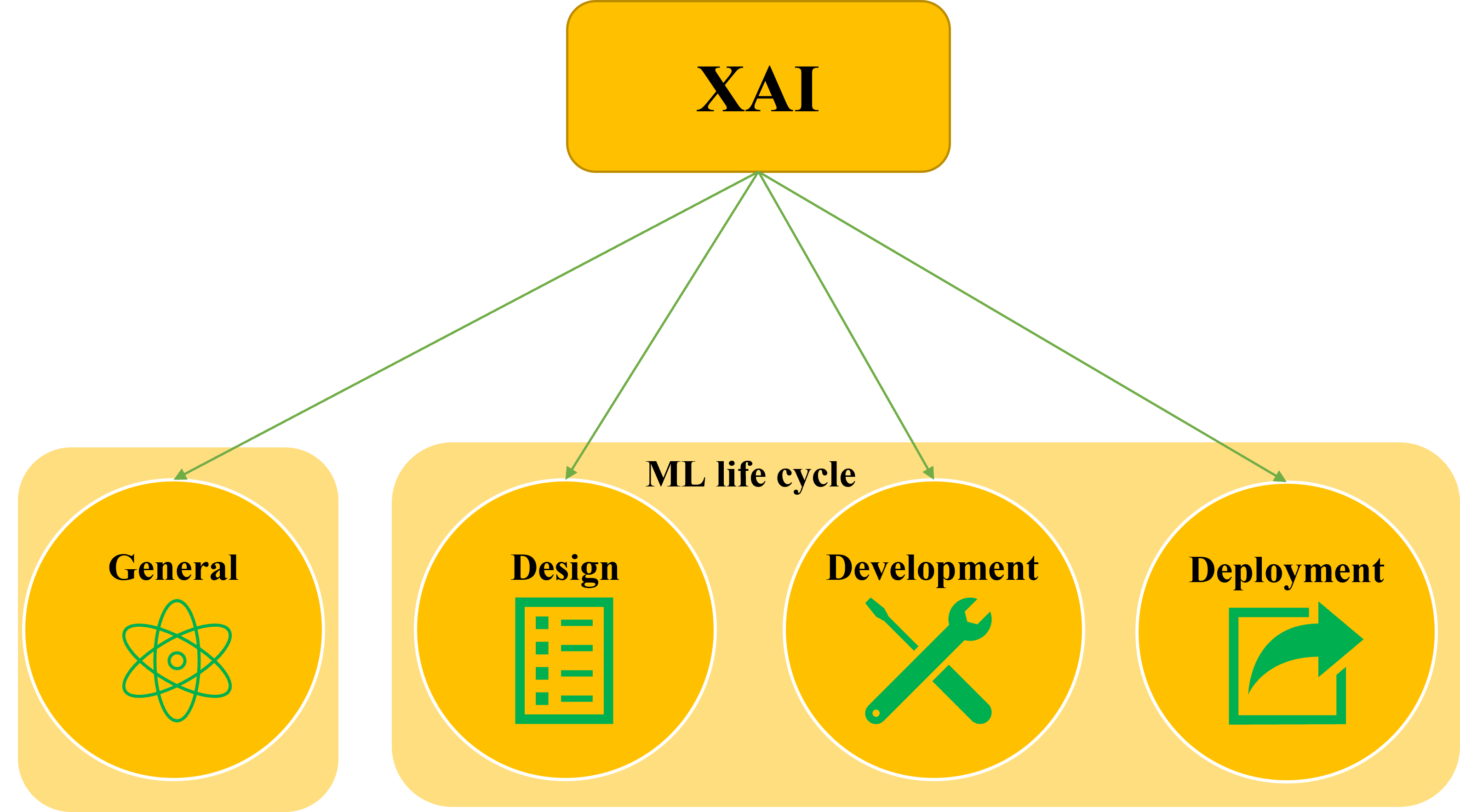}
\caption{The proposed organization to discuss the challenges and research directions in XAI. For simplicity, the arrows that show the flow in the life cycle are removed.}
\label{ml_cycle}
\end{figure}

This section discusses the challenges and research directions in XAI. In order to place them in a meaningful context, the discussion is organized into two main themes, as shown in Fig. \ref{ml_cycle}. The first theme focuses on the general challenges and research directions in XAI. The second theme is about the challenges and research directions of XAI based on the ML life cycle's phases. For simplicity, we divided the life cycle into three main phases: design, development, and deployment phases. The following subsections are shed light on these challenges and research directions.

\begin{table}
%\centering
\caption{A summary of the selected papers, categorized by phases as well as challenges and research directions.\label{challenges}}
\begin{tabular}{lp{8.7cm}p{5.3cm}}
\toprule
Phases & Challenges and research directions & Papers\\
\midrule

\multirow{2}{*}{General} & 
\hyperlink{Towards_more_formalism}{Towards more formalism} & \cite{Adadi2018,Dosilovic2018-xu,Guidotti2018-tx,Molnar2020,Payrovnaziri2020-xi,Carvalho,Aurangzeb,arrieta2020,Reyes,Samek2019-po,Longo2020,Pocevi020,Mi2020,Burkart2021-tq,Li2020,nunes2017systematic,seeliger2019semantic}, \cite{xie2020explainable,lucieri2020achievements,mueller2019explanation,buhrmester2019analysis,islam2021explainable,Messina} $^*$\\
& \hyperlink{Multidisciplinary}{Multidisciplinary research collaborations} & \cite{MILLER20191,Longo2020,Burkart2021-tq,Kovalerchuk2021,Fan2021,Abdul2018}, \cite{islam2021explainable,vilone2020explainable,lucieri2020achievements} *\\
& \hyperlink{expertise}{Explanations and the nature of user experience and expertise} & \cite{gunning2019xai,Chatzimparmpas,Naiseh2020,Aurangzeb,nunes2017systematic,arrieta2020}, \cite{mueller2019explanation,xie2020explainable,lucieri2020achievements} $^*$\\
& \hyperlink{trustworthiness}{XAI for trustworthiness AI}&  \cite{Pocevi020,Ras2018,Markus2021-bp,Chakraborty,islam2021explainable,belle2021principles, Chatzimparmpas,arrieta2020},\cite{xie2020explainable}*\\
& \hyperlink{trade-off}{Interpretability vs. performance trade-off}& 
\cite{arrieta2020,gunning2019xai,Aurangzeb,danilevsky2020survey,Longo2020}\\
& \hyperlink{Causal}{Causal explanations} & \cite{Molnar2020,Aurangzeb,belle2021principles,Pocevi020,Moraffah2020}\\
& \hyperlink{Contrastive}{Contrastive and counterfactual explanations} & \cite{Stepin2021}, \cite{mueller2019explanation} $^*$\\
& \hyperlink{heterogeneous}{XAI for non-image, non-text, and heterogeneous data} & \cite{Reyes,Kovalerchuk2021}, \cite{yuan2020explainability,lucieri2020achievements}*\\
& \hyperlink{composition}{Explainability methods composition} & \cite{Aurangzeb,Adadi2018,belle2021principles,Ras2018}, \cite{lucieri2020achievements} $^*$\\
& \hyperlink{models/methods}{Challenges in the existing XAI models/methods} & \cite{Aurangzeb,Molnar2020,Samek2019-po}\\
& \hyperlink{generation}{Natural language generation} & \cite{lucieri2020achievements}*\\
& \hyperlink{models}{Analyzing models, not data}& \cite{Molnar2021-do}\\
& \hyperlink{uncertainties}{Communicating uncertainties} & \cite{Chatzimparmpas,Molnar2020}\\
& \hyperlink{Time}{Time constraints}
& \cite{xie2020explainable,Doshi-Velez2017-yb}*\\
& \hyperlink{Reproducibility}{Reproducibility} & \cite{Payrovnaziri2020-xi}\\
& \hyperlink{economics}{The economics of explanations} &  \cite{Adadi2018}\\

\\

\multirow{2}{*}{Design} & \hyperlink{data_quality}{Communicating data quality} & \cite{Markus2021-bp,Burkart2021-tq} \\ %royal2019
& \hyperlink{sharing}{Data sharing} &  \cite{Holzinger}\\ 
\\

\multirow{2}{*}{Development} & \hyperlink{infusion}{Knowledge infusion} & \cite{Choo8402187,Zhang2018,Li2020}, \cite{Messina,zhang2020survey}$^*$ \\
& \hyperlink{training_process}{Developing approaches supporting explaining the training process} & \cite{Choo8402187,Chatzimparmpas,gunning2019xai}\\
& \hyperlink{debugging}{Developing model debugging techniques} &  \cite{Zhang2018}, \cite{xie2020explainable}$^*$ \\
& \hyperlink{comparison}{Using interpretability/explainability for models/architectures comparison} & \cite{Chatzimparmpas}\\
& \hyperlink{visual_analytics}{Developing visual analytics approaches for advanced DL architectures} & \cite{Chatzimparmpas,Choo8402187}\\
& \hyperlink{sparsity}{Sparsity of analysis} & \cite{dao2020demystifying}*\\
& \hyperlink{innovation}{Model innovation} &\cite{LIANG2021168,belle2021principles}\\
& \hyperlink{rules}{Rules extraction} & \cite{HE2020346,Abdul2018,Fan2021}\\
& \hyperlink{bayesian}{Bayesian approach to interpretability} & \cite{Chakraborty}\\
& \hyperlink{competencies}{Explaining competencies} & \cite{gunning2019xai}\\
\\

\multirow{2}{*}{Deployment} & \hyperlink{ontologies}{Improving explanations with ontologies} & \cite{Burkart2021-tq} \\
& \hyperlink{privacy}{XAI and privacy} & \cite{Longo2020,Holzinger} \\
& \hyperlink{security}{XAI and security} & \cite{arrieta2020,Tjoa2020,LIANG2021168,Ras2018}\\
& \hyperlink{safety}{XAI and safety} & \cite{HUANG2020100270,NaisehJiang2020,arrieta2020}\\
& \hyperlink{teaming}{Human-machine teaming} & \cite{seeliger2019semantic,Ras2018,gunning2019xai,NaisehJiang2020,Adadi2018,Abdul2018,Choo8402187,Chatzimparmpas}, \cite{islam2021explainable,Messina}* \\
& \hyperlink{agency}{Explainable agency} & \cite{Anjomshoae}\\
& \hyperlink{mexplanation}{Machine-to-machine explanation} & \cite{Preece2018,Weller2019,Adadi2018}\\
& \hyperlink{reinforcement}{XAI and reinforcement learning} & \cite{Lipton2018-dx,Guidotti2018-tx,Wells}, \cite{dao2020demystifying}*\\
& \hyperlink{planning}{Explainable AI planning (XAIP)} & \cite{Adadi2018},\cite{fox2017explainable}*\\
& \hyperlink{recommendation}{Explainable recommendation} & \cite{ZhangINR-066,NaisehJiang2020}\\
& \hyperlink{service}{XAI as a service}& \cite{Molnar2021-do}\\

\bottomrule
\end{tabular}
\\
\footnotesize{$^*$ A non-peer-reviewed paper from arXiv.}
\end{table}

\subsection{General Challenges and Research Directions in XAI}
In this section, we reported the general challenges and research directions in XAI.

\subsubsection{Towards more formalism}
\hypertarget{Towards_more_formalism}
It is one of the most raised challenges in the literature of XAI \cite{Adadi2018,Dosilovic2018-xu,Guidotti2018-tx,Molnar2020,Payrovnaziri2020-xi,xie2020explainable,Carvalho,Aurangzeb,mueller2019explanation,lucieri2020achievements,arrieta2020,Reyes,Samek2019-po,buhrmester2019analysis,Longo2020,Pocevi020,Mi2020,Burkart2021-tq,islam2021explainable,Li2020,nunes2017systematic,seeliger2019semantic,Messina}. It was suggested that more formalism should be considered in terms of systematic definitions, abstraction, and formalizing and quantifying \cite{Adadi2018}.

Starting from the need for systematic definitions, until now, there is no agreement on what an explanation is \cite{Guidotti2018-tx}. Furthermore, it has been found that similar or identical concepts are called by different names and different concepts are called by the same names \cite{Adadi2018,Dosilovic2018-xu}. In addition, without a satisfying definition of interpretability, how it is possible to determine if a new approach better explains ML models \cite{Molnar2020}? Therefore, to facilitate easier sharing of results and information, definitions must be agreed upon \cite{Adadi2018, Dosilovic2018-xu}.

With regards to the abstraction, many works have been proposed in an isolated way; thus, there is a need to be consolidated to build generic explainable frameworks that would guide the development of end-to-end explainable approaches \cite{Adadi2018}. Additionally, taking advantage of the abstraction explanations in identifying properties and generating hypotheses about data-generating processes (e.g., causal relationships) could be essential for future artificial general intelligence (AGI) systems \cite{Dosilovic2018-xu}.

Regarding the formalization and quantification of explanations, it was highlighted in \cite{Adadi2018} that some current works focus on a detailed problem formulation which becomes irrelevant as the method of interpretation or the explanation differs. Therefore, regardless of components that may differ, the expansibility problem must be generalized and formulated rigorously, and this will improve the state-of-the-art for identifying, classifying, and evaluating sub-issues of explainability \cite{Adadi2018}.

Establishing formalized rigorous evaluation metrics need to be considered as well \cite{Adadi2018}. However, due to the absence of an agreement on the definitions of interpretability/explainability, no established approach exists to evaluating XAI results \cite{Payrovnaziri2020-xi}. The lack of ground truth in most cases is the biggest challenge for rigorous evaluations \cite{xie2020explainable,Molnar2020}. So far, different evaluation metrics have been proposed, such as reliability, trustworthiness, usefulness, soundness, completeness, compactness, comprehensibility, human-friendly or human-centered, correctness or fidelity, complexity, generalizability \cite{xie2020explainable}. However, it seems that there are two main evaluation metrics groups: objective and human-centered evaluations \cite{Molnar2020}. The former is quantifiable mathematical metrics, and the latter relies on user studies \cite{Molnar2020}. Further progress are needed towards evaluating XAI techniques' performance and establishing objective metrics for evaluating XAI approaches in different contexts, models, and applications \cite{arrieta2020}. Achieving that may help in developing a model-agnostic framework that can suggest the most appropriate explanation taking into account problem domain, use case, and user's type \cite{Carvalho}.

\subsubsection{Multidisciplinary research collaborations}
\hypertarget{Multidisciplinary}
One area of research that can offer new insights for explainable methods is working closely with researchers from other disciplines such as psychology, behavioral and social sciences, human-computer interaction, physics, and neuroscience. Multidisciplinary research is therefore imperative to promote human-centric AI and expand utilizing XAI in critical applications \cite{islam2021explainable}.

Several studies, for instance \cite{Longo2020,Burkart2021-tq,lucieri2020achievements,Kovalerchuk2021,Fan2021,MILLER20191,Abdul2018}, have been suggested some potential multidisciplinary research works. In \cite{Longo2020}, it has been highlighted that approaching psychology discipline can help to get insights on both the structure and the attributes of explanations and the way they can influence humans. They also have suggested that defining the context of explanations is an important research direction. Here, it is essential to consider the domain of application, the users, type of explanations (e.g., textual, visual, combinations of solutions), and how to provide the explanations to the users. This research direction can form a connection with behavioral and social sciences. The paper in \cite{MILLER20191} also has shown that XAI can benefit from the work in philosophy, cognitive psychology/science, and social psychology. The paper summarizes some findings and suggests ways to incorporate these findings into work on XAI.

Approaching HCI studies are essential to XAI. However, few user experiments have been conducted in the area of explainability \cite{Burkart2021-tq}. Therefore, more should be conducted to study the topic adequately \cite{Burkart2021-tq}. Humans must be included in the process of creating and utilizing XAI models, as well as enhancing their interpretability/explainability \cite{Longo2020}. In \cite{lucieri2020achievements}, it has been highlighted that interactive tools may help users understand, test, and engage with AI algorithms, thereby developing new approaches that can improve algorithms' explainability. Furthermore, interactive techniques can help users to interpret predictions and hypothesis-test users' intuitions rather than relying solely upon algorithms to explain things for them. In \cite{Abdul2018}, it has been suggested drawing from the HCI research on interaction design and software learnability to improve the usability of intelligible or explainable interfaces. Additionally, HCI researchers can take advantage of the theoretical work on the cognitive psychology of explanations to make understandable explanations. They can also empirically evaluate the effectiveness of new explanation interfaces.

The advances in neuroscience should be of great benefit to the development and interpretation of DL techniques (e.g., cost function, optimization algorithm, and bio-plausible architectural design) owing to the close relationship between biological and neural networks \cite{Fan2021}. It is imperative to learn from biological neural networks so that better and explainable neural network architectures can be designed \cite{Fan2021}. Finally, connecting with physics and other disciplines that have a history of explainable visual methods might provide new insights for explainable methods \cite{Kovalerchuk2021}.

\subsubsection{Explanations and the nature of user experience and expertise}
\hypertarget{expertise}
Based on the nature of the application, users who use ML models can vary (e.g., data scientists, domain experts, decision-makers, and non-experts). The nature of user experience and expertise matters in terms of what kind of cognitive chunks they possess and the complexity they expect in their explanations \cite{Doshi-Velez2017-yb}. In general, users have varying backgrounds, knowledge, and communication styles \cite{Doshi-Velez2017-yb}. However, it seems that the current focus of explanation methods is tailored to users who can interpret the explanations based on their knowledge in the ML process \cite{Ras2018,xie2020explainable}.

The works in \cite{xie2020explainable,mueller2019explanation,gunning2019xai,Chatzimparmpas,Naiseh2020,Aurangzeb,nunes2017systematic} have highlighted what is needed to be considered with regards to explanations and the nature of user experience and expertise. In \cite{xie2020explainable}, user-friendly explanations have been suggested so users can interpret the explanations with less technical knowledge. Therefore, figuring out what to explain should follow the identification of the end-user. In \cite{mueller2019explanation}, it has been highlighted that previous works in explainable AI systems (e.g., expert systems) generally neglected to take into account the knowledge, goals, skills, and abilities of users. Additionally, the goals of users, systems, and explanations were not clearly defined. Therefore, clearly stating goals and purposes are needed to foster explanations testing within the appropriate context. In \cite{Naiseh2020}, the authors have discussed that identifying the users' goals and keeping up with their dynamic nature means collecting more data from them. It is also essential to develop changes detection approaches of goals and needs for the purpose of adapting these changes to end-users. For a deeper understanding of these dynamics, user studies (e.g., diary studies, interviews, and observation) can help develop guidelines for developing long-term explainable systems and determining which user data to gather to improve personalization.

In \cite{gunning2019xai}, it has been suggested that abstraction can be used to simplify the explanations. Understanding how abstractions are discovered and shared in learning and explanation is an essential part of the current XAI research. The work in \cite{Aurangzeb} has mentioned that the inclusion of end-users in the design of black-box AI models is essential, especially for specific domains, e.g., the medical domain. That would help to understand better how the end-users will use the outputs and interpret explanations. It is a way to educate them about the predictions and explanations produced by the system. In \cite{Chatzimparmpas}, the authors have discussed that utilizing users' previous knowledge is a significant challenge for visualization tools today. Customizing visualization tools for different user types can be useful at several stages of the ML model pipeline. However, to use prior users' knowledge in predictive models, it is important to establish processes to digitally capture and quantify their knowledge. 

In \cite{lucieri2020achievements}, it has been mentioned that DL models often use concepts that are unintelligible to predict outcomes. Therefore, using systems that use such models requires human-centric explanations that can accurately explain a decision and make sense to the users (e.g., medical domain expert). An approach to come with human-centric explanations is examining the role of human-understandable concepts acquired by DL models. It is also essential to analyze the features used by the DL models in predicting correct decisions, but based on incorrect reasoning. Having an understanding of the model’s concepts would help reduce reliability concerns and develop trust when deploying the system, especially in critical applications. The authors also highlighted the importance of addressing the domain-specific needs of specific applications and their users when developing XAI methods. Finally, the work in \cite{arrieta2020} has discussed that XAI can facilitate the process of explaining to non-experts how a model reached a given decision, which can substantially increase information exchange among heterogeneous people regarding the knowledge learned by models, especially when working in projects with multi-disciplinary team.

To sum up, it is crucial to tailor explanations based on user experience and expertise. Explanations should be provided differently to different users in different contexts \cite{royal2019}. In addition, it is also essential to clearly define the goals of users, systems, and explanations. Stakeholder engagement and system design are both required to understand which explanation type is needed \cite{royal2019}.

\setcounter{footnote}{0} 
\subsubsection{XAI for trustworthiness AI} 
\hypertarget{trustworthiness}
Increasing the use of AI in everyday life applications will increase the need for AI trustworthiness, especially in situations where undesirable decisions may have severe consequences \cite{xie2020explainable}. The High-Level Expert Group in European Commission put seven essentials for achieving trustworthy AI \footnote{\url{https://ec.europa.eu/commission/presscorner/detail/en/IP\_19\_1893}}: (1)\emph{human agency and oversight}; (2) \emph{robustness and safety}; (3) \emph{privacy and data governance}; (4) \emph{transparency}; (5) \emph{diversity, non-discrimination and fairness}; (6) \emph{societal and environmental well-being}; and (7) \emph{accountability}. The discussion about privacy, security, and safety are given in \hyperlink{privacy}{XAI and Privacy} Section, \hyperlink{security}{XAI and Security} Section, and \hyperlink{safety}{XAI and Safety} Section, respectively. The discussion in this section is for what is reported in the selected papers regarding fairness and accountability.

With regards to fairness, ML algorithms must not be biased or discriminatory in the decisions they provide. However, with the increased usage of ML techniques, new ethical, policy, and legal challenges have also emerged, for example, the risk of unintentionally encoding bias into ML decisions \cite{Chakraborty}. Meanwhile, the opaque nature of data mining processes and the complexity of ML make it more challenging to justify consequential decisions \cite{Chakraborty}. The work in \cite{islam2021explainable} argues that data, algorithmic, and social biases need to be remedied in order to promote fairness. Further, it is imperative to be able to analyze AI systems to have trust in the model and its predictions, especially for some critical applications. Researchers started trying to form a definition of fairness and the meaning of fairness in an algorithm as discussed in \cite{Chakraborty}. According to \cite{Chakraborty}, it would also be necessary to devise new techniques for discrimination-aware data mining. It is also worth noting that when converting fairness into a computational problem, we need to keep the fairness measures fair \cite{Chakraborty}. The work in \cite{Ras2018} states that it is possible to visualize learned features using XAI methods and assess bias using methods other than explanation methods. On the other hand, regulations and laws are necessary for the suspicion about unfair outcomes \cite{Ras2018}.

Having accountability means having someone responsible for the results of AI decisions if harm occurs. In \cite{royal2019}, it has been mentioned that investigating and appealing decisions with major consequences for people is an important aspect of systems of accountability, and some current regulations also aim to achieve this. XAI can be an important factor in systems of accountability by providing users with the means to appeal a decision or modify their behavior in the future to achieve a better result. However, more work should be done to establish an environment that promotes individual autonomy and establish a system of accountability. It has also been discussed in \cite{Chakraborty} that developing procedures for testing AI algorithms for policy compliance is necessary so that we can establish whether or not a given algorithm adheres to a specific policy without revealing its proprietary information. It is also desirable for a model to specify its purposes and provide external verification of whether these goals are met and, if not, describe the causes of the predicted outcomes.

The use of XAI can enhance understanding, increase trust, and uncover potential risks \cite{islam2021explainable}. Therefore, when designing XAI techniques, it is imperative to maintain fairness, accountability, and transparency \cite{islam2021explainable}. On the other hand, it is necessary to highlight that not only black-box AI models are vulnerable to adversarial attacks, but also XAI approaches \cite{slack2020fooling}. There is also a risk that to promote trust in black-box AI models predictions; explainers may be more persuasive but misleading than informative, so users may become deceived, thinking the system to be trustworthy \cite{Pocevi020,royal2019}. It is possible to increase trust through explanations, but explanations do not always produce systems that produce trustworthy outputs or ensure that system implementers make trustworthy claims about its abilities \cite{royal2019}.

The work in \cite{Markus2021-bp} discusses measures to create trustworthy AI. It has been highlighted that before employing AI systems in practice, it is essential to have quantitative proxy metrics to assess explanation quality objectively, compare explanation methods, and complement them with human evaluation methods (e.g., data quality reporting, extensive testing, and regulation).

Finally, it is good to note that a further explore the idea of Responsible AI with a discussion about principles of AI, fairness, privacy, and data fusion can be found in \cite{arrieta2020}.

\subsubsection{Interpretability vs. performance trade-off}
\hypertarget{trade-off}
The belief that complicated models provide more accurate outcomes is not necessarily correct \cite{rudin2019stop}. However, this can be incorrect in cases when the given data is structured and with meaningful features \cite{rudin2019stop}. In a situation where the function being approximated is complex, that the given data is widely distributed among suitable values for each variable and the given data is sufficient to generate a complex model, the statement \emph{``models that are more complex are more accurate''} can be true \cite{arrieta2020}. In such a situation, the trade-off between interpretability and performance becomes apparent \cite{arrieta2020}.

When the performance is coupled with model complexity, model interpretability is in question \cite{arrieta2020}. Explainability techniques, however, could help in minimizing the trade-off \cite{arrieta2020}. However, according to \cite{Aurangzeb}, what determines this trade-off? and who determines it? The authors have highlighted the importance of discussing with end-users this trade-off so that they can be aware of the potential risks of misclassification or opacity. Another point that should be considered is the approximation dilemma: models need to be explained in enough detail and in a way that matches the audience for whom they are intended while keeping in mind that explanations reflect the model and do not oversimplify its essential features \cite{arrieta2020}. Even though studying the trade-off is essential, it is impossible to proceed without standardized metrics for assessing the quality of explanations \cite{danilevsky2020survey}.

Another possible solution for the trade-off is suggested in \cite{Longo2020} which is developing fully transparent models throughout the entire process of creation, exploitation, and exploration and can provide local and global explanations. In turn, this leads to using methods that embed learning capabilities to develop accurate models and representations. The methods should also be able to describe these representations in effective natural language consistent with human understanding and reasoning.

\subsubsection{Causal explanations}
\hypertarget{Causal}
Developing causal explanations for AI algorithms (i.e., why they made those predictions instead of how they arrived at those predictions) can help increasing human understanding \cite{Pocevi020}. In addition, causal explanations strengthen models' resistance to adversarial attacks, and they gain more value when they become part  of decision-making \cite{Molnar2020}. However, there can be conflicts between predicting performance and causality \cite{Molnar2020}. For example, when the confounder, which is a variable that influences both the dependent variable and independent variable, is missing from the model \cite{Molnar2020}.

Causal explanations are anticipated to be the next frontier of ML research and to become an essential part of the XAI literature \cite{Aurangzeb,belle2021principles}. There is a need for further research to determine when causal explanations can be made from an ML model \cite{Molnar2020}. In addition, according to a recent survey on causal interpretability for ML \cite{Moraffah2020}, it has been highlighted the absence of ground truth data for causal explanations and verification of causal relationships make evaluating causal interpretability more challenging. Therefore, more research is needed to guide on how to evaluate causal interpretability models \cite{Moraffah2020}.

\subsubsection{Contrastive and counterfactual explanations}
\hypertarget{Contrastive}
Contrastive explanations describe why one event occurred but not another, while counterfactual explanations describe what is needed to produce a contrastive output with minimal changes in the input \cite{Stepin2021}. Questions in the contrastive form "Why x and not y?" and questions of the counterfactual form "What if?" and "What would happen if?" \cite{mueller2019explanation}.

In a recent survey of contrastive and counterfactual explanations \cite{Stepin2021}, it has been found that contrastive and counterfactual explanations help improve the interaction between humans and machines and personalize the explanation of algorithms. A further important point as observed by \cite{Stepin2021} that one of the significant barriers towards a fair assessment of new frameworks is the lack of standardization of evaluation methods. The theoretical frameworks are also found inadequate for applying to XAI as a result of the disconnect between the philosophical accounts of counterfactual explanation to scientific modeling as well as ML-related concepts. Furthermore, it has been found that different domains of science define counterfactual explanations differently, as do the approaches used to solve specific tasks.

In the light of possible research directions in this point, it has been suggested in \cite{Stepin2021} the importance of including end-users in the evaluation of generated explanations since these explanations are designed to be user-oriented. In addition, since contrastive and counterfactual explanations address causal and non-causal relationships, new horizons open to the XAI community by unifying causal and non-causal explanatory engines within a contfactually-driven framework. Furthermore, bringing together researchers from the humanities and the computational sciences could contribute to further development for contrastive and counterfactual explanations generation.

\subsubsection{XAI for non-image, non-text, and heterogeneous data}
\hypertarget{heterogeneous}
The focus of XAI works is mainly on image and text data. However, other data types exist but until now not well explained, such as sequences, graphs, and Spatio-temporal data.

Using visualization to transform non-image data into images creates opportunities to discover explanations through salient pixels and features \cite{Kovalerchuk2021}. However, this should not be the only way for explainability for non-image or non-text data. For example, existing explanation approaches for image or text data need to be adjusted to be used with graph data (without any transformation) \cite{yuan2020explainability}. Additionally, there is a need to develop new approaches for explaining the information that exists with non-image or non-text data, e.g., structural information for graph data \cite{yuan2020explainability}.

Finally, with the advent of AI systems that use various types of data, explainability approaches that can handle such heterogeneity of information are more promising \cite{Reyes}. For example, such systems can simulate clinicians' diagnostic processes in the medical domain where both images and physical parameters are utilized to make decisions \cite{lucieri2020achievements}. Thus, they can enhance the diagnostic effectiveness of the systems as well as explain phenomena more thoroughly \cite{lucieri2020achievements}.

\subsubsection{Explainability methods composition}
\hypertarget{composition}
For specific applications in healthcare (e.g., predicting disease progression), several types of explanations at different levels are needed  (e.g., local and global explanations) \cite{Aurangzeb} in order to provide the most complete and diverse explanations we can \cite{lucieri2020achievements}. This is derived from the way clinicians communicate decisions utilizing visualizations and temporal coherence as well as textual descriptions \cite{lucieri2020achievements}.

Some overlap exists between explainability methods, but for the most part, each seems to address a different question \cite{belle2021principles}. According to \cite{Adadi2018}, combining various methods to obtain more powerful explanations is rarely considered. In addition, rather than using disparate methods separately, we should investigate how we can use them as basic components that can be linked and synergized to develop innovative technologies \cite{Adadi2018}. It is argued that enabling composability in XAI may contribute to enhancing both explainability and accuracy \cite{Adadi2018}. Furthermore, it could help to provide answers in a simple human interpretable language \cite{Ras2018}. First efforts, as cited in \cite{lucieri2020achievements}, have been recently made as in \cite{Park_2018_CVPR} where the authors proposed a model that can provide visual relevance and textual explanations.

\subsubsection{Challenges in the existing XAI models/methods}
\hypertarget{models/methods}
There are some challenges in the existing XAI models/methods that have been discussed in the literature. Starting with scalability, which is a challenge that exists in explainable models as discussed in \cite{Aurangzeb}. For example, each case requiring an explanation entails creating a local model using LIME explainable model \cite{ribeiro2016should}. The scalability can be an issue when there is a huge number of cases for which prediction and explanation are needed. Likewise, when computing Shapley values \cite{vstrumbelj2014explaining}, all combinations of variables must be considered when computing variable contributions. Therefore, such computations can be costly for problems that have lots of variables.

Feature dependence presents problems in attribution and extrapolation \cite{Molnar2020}. If features are correlated, attribution of importance and features effects becomes challenging. For sensitivity analyses that permute features, when the permuted feature has some dependence on another feature, the association breaks, resulting in data points outside the distribution, which could cause misleading explanations. 

In \cite{Samek2019-po}, the authors discussed some limitations with heatmaps explanations. Heatmaps explanations visualize what features are relevant for making predictions. However, the relation between these features, either individually or in combination, remains unclear. Low abstraction levels of explanations are another limitation. Heatmaps highlight that specific pixels are significant without indicating how the relevance values relate to abstract concepts in the image, such as objects or scenes. The model's behavior can be explained in more abstract, more easily understood ways by meta-explanations that combine evidence from low-level heatmaps. Therefore, further research is needed on meta-explanations.

Model-based (i.e., ante-hoc models) and post-hoc explainability models have some challenges, as have been discussed in \cite{Murdoch22071}. When model-based methods cannot predict with reasonable accuracy, practitioners start the search for more accurate models. Therefore, one way to increase the usage of model-based methods is to develop new modeling methods that maintain the model's interpretability and render more accurate predictions. More details about this direction are provided in \cite{rudin2019stop}. Further, for model-based methods, there is a need to develop more tools for feature engineering. It is possible to achieve comparable predictive accuracy in some applications when the features are more informative and meaningful. Therefore, enhancing the possibility of model-based methods can be accomplished by producing more useful features. Two categories of works can help achieve that: improve tools for exploratory data analysis and improve unsupervised techniques. The former helps to understand the data, and domain knowledge could help to identify helpful features. The latter is needed because unsupervised techniques are often used to identify relevant structures automatically, so advances in unsupervised techniques may result in better features.

The authors in \cite{Murdoch22071} have also discussed some challenges for post-hoc explainability models. According to the authors, it is challenging to determine what format or combination of formats will adequately describe the model's behavior. Furthermore, there is uncertainty over whether the current explanation methods are adequate to capture a model’s behavior or novel methods still needed. Another challenge is if post-hoc explanations methods identify learned relationships by the model that practitioners know to be incorrect, is it possible that practitioners fix these relationships learned and increase the predictive accuracy? Further research in pos-thoc explanations can help exploit prior knowledge to improve the predictive accuracy of the models.

\subsubsection{Natural language generation}
\hypertarget{generation}
Explaining in natural language needs to be accurate, useful, and easy to understand \cite{reiter-2019-natural}. Furthermore, in order to produce good quality explanations, the generated explanations need to be tailored to a specific purpose and audience, be narrative and structured, and communicate uncertainty and data quality that could affect the system's output \cite{reiter-2019-natural}.

Four challenges that are crucial in generating good quality explanations have been discussed in \cite{reiter-2019-natural}:
\begin{itemize}
    \item Evaluation challenge: Develop inexpensive but reliable ways of estimating scrutability, trust, etc. Do we have a chance to obtain reliable results if we ask users to read explanations and estimate, for example, scrupability? What experimental design gives the best results? Before we do these steps, should we make sure the explanations are accurate?

    \item Vague Language challenge: Using vague terms in explanations is much easier to understand by humans because they think in qualitative terms \cite{van2012not}. However, how can vague language be used in explanations, such that the user does not interpret it in a way that will lead to a misunderstanding of the situation? In addition, setting the priority of messages based upon features and concepts that the user is aware of would be helpful. Furthermore, phrasing and terminology used should be intuitive to users.
    
    \item Narrative challenge: Explaining symbolic reasoning narratively is more straightforward to comprehend than numbers and probabilities \cite{daniel2017thinking}. Therefore, we need to develop algorithms for creating narrative explanations to present the reasoning.

    \item Communicating data quality challenge: Techniques should be developed to keep users informed when data problems affect results. We have discussed this issue in detail in \hyperlink{data_quality}{Communicating Data Quality} Section.
\end{itemize}

Another challenge has been discussed in \cite{lucieri2020achievements}. In some medical domains, it could be necessary for AI systems to generate long textual coherent reports to mimic the behavior of doctors. The challenge here is that after generating a few coherent sentences, language generation models usually start producing seemingly random words that have no connection to previously generated words. One of the solutions to this problem would be to use transformer networks \cite{vaswani2017attention} as language model decoders, which can capture word relationships in a longer sentence. In order to evaluate the generated reports, it is essential to compare them with human-generated reports. However, since human-generated reports are usually free-text reports (i.e., not following any specific template), it is important to first eliminate unnecessary information for the final diagnosis from human-generated reports then conduct the comparison.

\subsubsection{Analyzing models, not data}
\hypertarget{models}
The author in \cite{Molnar2021-do} has discussed that analyzing models instead of data is a possible future of ML interpretability. It has been mentioned that a possible way to extract knowledge from data is through interpretable ML. That is because an ML model can automatically identify if and how features are important to predicting outcomes and recognize how relationships are represented.

He added that there is a need to move from analyzing assumption-based data models to analyzing assumption-free black-box AI models. That's because making assumptions about the data (i.e., distribution assumptions) is problematic. Typically, they are wrong (do not follow the Gaussian distribution), hard to check, extremely inflexible, and difficult to automate. Further, assumption-based data models in many domains are typically less predictive than black-box AI models (i.e., generalization) when having lots of data, which is available due to digitization. Therefore, the author has argued that there should be a development of all the tools that statistics offer for answering questions (e.g., hypothesis tests, correlation measures, interaction measures) and rewrite them for black-box AI models. To some extent, this is already taking place. For example, in a linear model, the coefficients quantify the effects of an individual feature on the result. The partial dependent plot \cite{friedman2001greedy} represents this idea in a more generalized form.

\subsubsection{Communicating uncertainties}
\hypertarget{uncertainties}
Communicating uncertainty is an important research direction because it can help to inform the users about the underlying uncertainties in the model and explanations. According to \cite{Chatzimparmpas}, there are already inherent uncertainties in ML models; and model refinement efforts by developers may introduce new uncertainties (e.g., overfitting). Furthermore, some explanation methods such as permutation feature importance and Shapley value give explanations without measuring the uncertainty implied by the explanations \cite{Molnar2020}.

Quantifying uncertainty is an open research topic \cite{Chatzimparmpas}. However, some works exist towards quantifying uncertainty as discussed in \cite{Molnar2020}. The uncertainty surrounding ML models can take many forms and occur throughout the ML life phases \cite{Chatzimparmpas}. Therefore, in order to make progress, it is needed to become more rigorous in studying and reliably quantifying uncertainties at model's various phases and with the explanation methods and communicate them to the users, then users can respond accordingly \cite{Chatzimparmpas,Molnar2020}.

\subsubsection{Time constraints}
\hypertarget{Time}
Time is an essential factor in producing explanations and in interpretation. Some explanations must be produced promptly to let the user react to the decision \cite{xie2020explainable}. Producing explanations efficiently can save computing resources, thereby making it useful for industrial use or in environments with limited computing capability \cite{xie2020explainable}. In some situations (e.g., plant operation application), the provided explanations need to be understood quickly to help the end-user to make a decision \cite{Doshi-Velez2017-yb}. On the other hand, in some situations (e.g., scientific applications), users would likely be willing to devote considerable time understanding the provided explanation \cite{Doshi-Velez2017-yb}. Therefore, time is an essential factor considering the situation, available resources, and end-users.

\subsubsection{Reproducibility}
\hypertarget{Reproducibility}
In a recent review of XAI models based on electronic health records, it has been found that research reproducibility was not stressed well in the reviewed literature, though it is paramount \cite{Payrovnaziri2020-xi}. In order to facilitate comparisons between new ideas and existing works, researchers should use open data, describe the methodology and infrastructure they used, and share their code \cite{Payrovnaziri2020-xi}. In addition, it has been suggested that publication venues should establish reproducibility standards that authors must follow as part of their publication process \cite{Payrovnaziri2020-xi}.

\subsubsection{The economics of explanations}
\hypertarget{economics}
Research into the economic perspective of XAI is sparse, but it is essential \cite{Adadi2018}. With the pressures of social and ethical concerns about trusting black-box AI models, XAI has the potential to drive a real business value \cite{Adadi2018}. XAI, however, comes at a cost \cite{doshi2017accountability}.

Recently, the work in \cite{beaudouin2020flexible} identified costs of explanations in seven main categories (1) costs of explanations design, (2) costs of creating and storing audit logs, (3) costs of trade secrets violation (e.g., the forced disclosure of source code), (4) costs of slowing down innovation (e.g., increasing time-to-market), (5) costs of reducing decisional flexibility if the future situation does not justify the previous explanation, (6) cost of conflict with security and privacy matters, and (7) costs of using less efficient models for their interpretability. Therefore, costs associated with algorithmic explanations should be incurred when the benefits of the explanations outweigh the costs \cite{doshi2017accountability}. 

Cost estimation is one of the issues that should be addressed by encouraging economic interpretations. Other issues include algorithms proprietary, revealing trade secrets, and predicting XAI market evolution \cite{Adadi2018}.

\subsection{Challenges and Research Directions of XAI in the Design Phase}
In this phase, the data is collected from at least one source. Then, the data preparation step is done to prepare the collected data for the training phase. By grouping what was collected from the selected papers, we identified two main challenges that need further research: communicating data quality and data sharing.

\subsubsection{Communicating data quality} 
\hypertarget{data_quality}
%\label{data_quality}
The provided explanations for the AI system or its outcomes depend on the data used to build the system. Data bias, data incompleteness, and data incorrectness are issues that affect the quality of the data. Training AI systems using low-quality data will be reflected in their outcomes \cite{reiter-2019-natural}. For example, an AI system developed for lung cancer risks prediction using data from Americans may not accurately estimate risks for a resident of Delhi due to the differences in polluted environments in which they are living at \cite{reiter-2019-natural}. So, what can be of high quality for a particular purpose can be of low quality for another \cite{Markus2021-bp}. Reducing system accuracy is not the only consequence of building an AI system using low-quality data; producing unfair decisions and degrading the explainability of the AI system are other possible consequences.

With this in mind, it has been suggested to be aware of how data was collected and any limitations associated with the collected data \cite{royal2019}. Further, it has been highlighted the importance of clarifying any data issues that can reduce accuracy when producing explanations \cite{reiter-2019-natural}. However, how can we communicate data quality to users to let them know how the results are influenced by data used.

In \cite{ahmad2019challenge}, the authors discussed several issues that arise when producing explanations for AI models that use imputation of missing data. They recommended disclaimers accompanied by the derived explanations and educating end-users about the risks involved of incorrect explanations. Even though it is good to come with appropriate disclaimers, we believe that future studies should be undertaken to develop a practical and measurable way to communicate data quality to users. Proposing dimensions of data quality could be the basis for that. We recommend starting with the following questions which is inspired from the work in \cite{dama2020}:

\begin{itemize}
    \item Which essential dimensions of data quality are wanted?
    \item What are the definitions of those dimensions? and how to measure them?
    \item How to deal with them to improve the AI model and hence its explanations?
    \item How to communicate them (and highlight any possible risks)?
\end{itemize}

According to \cite{Burkart2021-tq}, there is a variety of data quality dimensions such as completeness, accuracy, and consistency. For an extensive list of dimensions of data quality that occur in information systems, the reader may refer to the research paper in \cite{dama2020}. The fairness dimension can also be included, which may include demographic parity differences. It is essential to highlight that the way that can be used to communicate data quality can vary based on the type of users.

\subsubsection{Data sharing}
\hypertarget{sharing}
Data privacy and data security are two major issues concerning XAI. Since AI is used as a data-driven method and because any requested explanations depend on data used to build AI systems, two main aspects related to data should be considered: data sharing and data preservation. 

Data sharing in this context means making raw data available to be used by other partners \cite{Holzinger}. Data preservation is the retention of raw data, at least until we stop using the AI solution. The discussion here focuses on data sharing, which is related to the data collection and preparation phase. Challenges related to data preservation are discussed later in \hyperlink{privacy}{XAI and Privacy} Section.

Data sharing is a significant challenge in many data-driven solutions \cite{Holzinger}. Two common strategies are used for data sharing between the partners: share raw data directly or send it to a central analysis repository \cite{Holzinger}. At this phase, no explanations that may violate privacy are demanded because the solutions are still not developed yet. However, it is essential for users (e.g., patients) to feel confident that their data is secured and protected from unauthorized access and misuse, and any processes are limited to the part of data that they have consented to \cite{Holzinger}. According to \cite{Holzinger}, the implementation of watermarking or fingerprinting are typical reactive techniques used to deal with this issue. Watermarking techniques prove the authenticity and ownership of a dataset, while fingerprinting techniques help to identify the data leak because partners receive the same basic set but marked differently with their fingerprints \cite{Holzinger}.

Federated learning can be a possible solution to avoid raw data sharing. Federated learning allows building ML models using raw data distributed across multiple devices or servers \cite{mcmahan2017communication,Wahab}. As described in \cite{Wahab}, the training using federated learning starts by sending initial model parameters by the central server, which are obtained after a few training iterations, to a set of clients. Then, each client uses its resources to train an ML model locally on its own dataset using the shared parameters. Afterward, each client sends the server an updated version of the parameters. As a result of aggregating clients' parameters, the server creates a global model. As soon as the global model reaches a certain accuracy level, the training process is stopped. 

Even though the data never leaves the user’s device, increasing the number of clients involved in a collaborative model makes it more susceptible to inference attacks intended to infer sensitive information from training data \cite{Federated1,Wahab}. Possible research directions to deal with privacy challenges of federated learning have been discussed in \cite{Wahab} such as privacy-preserving security assurance, defining optimal bounds of noise ratio, and proposing granular and adaptive privacy solutions.

\subsection{Challenges and Research Directions of XAI in the Development Phase}
There are three main types of learning in ML: supervised, unsupervised, and reinforcement learning. In supervised learning, a learning algorithm is used to train an ML model to capture patterns in the training data that map inputs to outputs. With unsupervised learning, which is used when only the input data is available, an ML model is trained to describe or extract relationships in the training data. For reinforcement learning, an ML model is trained to make decisions in a dynamic environment to perform a task to maximize a reward function. In the following subsections, we discuss the challenges and research directions during developing ML models.

\subsubsection{Knowledge infusion}
\hypertarget{infusion}
A promising research direction is incorporating human domain knowledge into the learning process (e.g., to capture desired patterns in the data). According to \cite{Messina}, understanding how experts analyze images and which regions of the image are essential to reaching a decision could be helpful to come with novel model architectures that mimic that process. Furthermore, our explanations can be better interpretable and more informative if we use more domain/task-specific terms \cite{zhang2020survey}.

Recently, the work in \cite{XIE2021101985} highlights various ways of incorporating approaches for medical domain knowledge with DL models such as transfer learning, curriculum learning, decision level fusion, and feature level fusion. According to that survey, it was seen that with appropriate integrating methods, different kinds of domain knowledge could be utilized to improve the effectiveness of DL models. A review focused on knowledge-aware methods for XAI is given by \cite{Li2020}. Based on the knowledge source, two categories are identified: knowledge methods and knowledge-based methods. Unstructured data is used as a knowledge source in knowledge methods, while knowledge-based methods use structured knowledge to build explanations. According to \cite{Li2020}, when we use external domain knowledge, we are able to produce explanations that identify important features and why they matter. As concluded in that survey, many questions remain unanswered regarding utilizing external knowledge effectively. For instance, in a vast knowledge space, how can relevant knowledge be obtained or retrieved? To demonstrate this point, let us take the Human-in-the-loop approach as an example. Typically, a user has a wide range of knowledge in multiple domains; thus, the XAI system must ensure that the knowledge provided to the user is desirable.

Recent works in the knowledge that can be incorporated during training ML are given in \cite{Fe-Fei,Gaur9357868,Zhang2018}. In \cite{Fe-Fei}, a one-shot learning technique was presented for incorporating knowledge about object categories, which may be obtained from previously learned models, to predict new objects when very few examples are available from a given class. Another work in \cite{Gaur9357868} has shown how knowledge graph is integrated into DL using knowledge-infused learning and presented examples on how to utilize knowledge-infused learning towards interpretability and explainability in education and healthcare. The work in \cite{Zhang2018} has mentioned that the middle-to-end learning of neural networks with weak supervision via human-computer interaction is believed to be a fundamental research direction in the future.

Based on all that, it can be seen that using XAI to explain the outcomes of the models (e.g., pointing which regions of the image were used to reach the decision) can help to understand better what was learned from the incorporated human knowledge. Thus, it would help to adjust the way used in incorporating the knowledge or come with innovations in model architectures. Furthermore, it could be used to confirm whether a model follows the injected knowledge and rules, especially with critical applications, e.g., autonomous driving model \cite{Choo8402187}. Therefore, more research is needed to investigate how experts can interact with ML models to understand them and improve their abilities, which would be a promising direction in which XAI can contribute.

\subsubsection{Developing approaches supporting explaining the training process}
\hypertarget{training_process}
Training ML models, especially DL, is a lengthy process that usually takes hours to days to finish, mainly because of the large datasets used to train the models \cite{Choo8402187}. Therefore, researchers and practitioners have contributed to developing systems that could help steer the training process and develop better models. 

\setcounter{footnote}{0} 
Examples of progressive visual analytics systems are cited in \cite{Choo8402187}. For example, DeepEyes \cite{Pezzotti} is an example of a progressive visual analytics system that enables advanced analysis of DNN models during training. The system can identify stable layers, identify degenerated filters that are worthless, identify inputs that are not processed by any filter in the network, reasons on the size of a layer, and it helps to decide whether more layers are needed or eliminate unnecessary layers. DGMTracker is another example \cite{Liu2018} which is developed for better understanding and diagnosing the training process of deep generative models (DGMs). In addition, big tech companies such as Google and Amazon have developed toolkits to debug and improve the performance of ML models such as Tensor-Board\footnote{\url{https://www.tensorflow.org/tensorboard}} and SageMaker Debugger\footnote{\url{https://aws.amazon.com/sagemaker/debugger/}}.

Future studies to deal with this challenge are therefore recommended in order to develop XAI approaches supporting the online training monitoring to get insights that could help to steer the training process by the experts, which could help in developing better models and minimizing time and resources \cite{Chatzimparmpas,Choo8402187}.

\subsubsection{Developing model debugging techniques} 
\hypertarget{debugging}
The model is already trained at this stage, and we want to discover any problems that can limit its predictions. The debugging of ML models is paramount for promoting trust in the processes and predictions, which could result in creating new applications \cite{hall2019proposed,Zhang2018}, e.g., visual applications for CNN. A variety of debugging techniques exists, including model assertion, security audit, variants of residual analysis and residual explanation, and unit tests \cite{hall2019proposed}. According to \cite{xie2020explainable}, understanding what causes errors in the model can form the foundation for developing interpretable explanations. The next step is developing more model debugging techniques and combining them with explanatory techniques to provide insight into the model's behavior, enhance its performance, and promote trust \cite{hall2019proposed}.

\subsubsection{Using interpretability/explainability for models/architectures comparison}
\hypertarget{comparison}
It is widely known that the performance of ML models/architectures varies from one dataset/task to another \cite{Chatzimparmpas}. Usually, error performance metrics are used for the comparison to choose the suitable model/architecture for the given dataset/task and to decide how to combine models/architectures for better performance \cite{Chatzimparmpas}. However, even if the models may have the same performance, they can use different features to reach the decisions \cite{Samek2017-lb}. Therefore, the interpretability/explainability of models can be helpful for models/architectures comparison \cite{Samek2017-lb}. It could even be said that the better we understand models' behavior and why they fail in some situations, the more we can use those insights to enhance them \cite{Samek2017-lb}. In the future, it is expected that explanations will be an essential part of a more extensive optimization process to achieve some goals such as improving a model’s performance or reducing its complexity \cite{Samek2019-po}. Further, XAI can be utilized in models/architectures comparison.

\subsubsection{Developing visual analytics approaches for advanced DL architectures}
\hypertarget{visual_analytics}
While visual analytic approaches have been developed for basic DL architectures (e.g., CNNs and RNNs), advanced DL architectures have yet to be addressed in this way (e.g., ResNet \cite{he2016deep} and DenseNet \cite{huang2017densely}) \cite{Chatzimparmpas,Choo8402187}. Advanced DL architectures pose several challenges for visual analytic and information visualization communities due to their large number of layers, the complexity of network design for each layer, and the highly connected structure between layers \cite{Choo8402187}. Therefore, developing efficient visual analytics approaches for such architectures in order to increase their interpretability as well as the explainability of their results is needed \cite{Chatzimparmpas,Choo8402187}.

\subsubsection{Sparsity of analysis}
\hypertarget{sparsity}
Interpreting and validating the reasoning behind a neural network classifier requires examining the saliency maps of various samples from input data, which can be a challenging task if there are a vast number of samples \cite{dao2020demystifying}. Therefore, the number of visualizations that a user has to analyze should be as small as possible to reduce the sparsity of the analysis \cite{dao2020demystifying}. A way to achieve that can be developing novel methods to identify a meaningful subset of the entire dataset to interpret; then, by using this meaningful subset, it is needed to come up with an interpretation of the relationship between various samples and various subsets \cite{dao2020demystifying}.

\subsubsection{Model innovation}
\hypertarget{innovation}
By explaining DL models, we can gain a deeper understanding of their internal structure and can lead to the emergence of new models (e.g., ZFNet \cite{Zeiler}) \cite{LIANG2021168}. Therefore, in the future, the development of explanation methods for DL and new DL models are expected to complement each other \cite{LIANG2021168}.

Another research area is developing new hybrid models where the expressiveness of opaque models is combined with the apparent semantics of transparent models (e.g., combining a neural network with a linear regression) \cite{belle2021principles}. This research area can be helpful for bridging the gap between opaque and transparent models and could help in developing highly efficient explainable models \cite{belle2021principles}.

\subsubsection{Rules extraction}
\hypertarget{rules}
Historically, the need for explanations dates back to the early works in explaining expert systems and Bayesian networks \cite{biran2017explanation}. Rule extraction from ML models has been studied for a long time \cite{towell1993extracting,OMLIN199641,ANDREWS1995373}. However, there is still an increasing interest in utilizing rule extraction for explainability/interpretability \cite{Zilke,HE2020346}. Therefore, to discover methods that may work for explainability/interpretability, we should revisit the past research works \cite{Abdul2018}.

According to \cite{ANDREWS1995373,HE2020346}, there are three main approaches for rule extraction: (1) Decomposition approach based on the principle that the rules are extracted at the neuron level, such as visualizing a network's structure, (2) Pedagogical approach that extracts rules that map inputs directly to outputs regardless of their underlying structure, such as computing gradient, (3) Eclectics approach, which is the combination of both decompositional and pedagogical approaches.

Further research for rules extraction is needed, which has been discussed in \cite{HE2020346}. First, visualize neural networks' internal structure. Through visualizing each activated weight connection/neuron/filter from input to output, one can understand how the network works internally and produce the output from the input. Second, transform a complex neural network into an interpretable structure by pruning unimportant or aggregating connections with similar functions. By doing this, the overfitting issue can be reduced, and the model's structure becomes easier to interpret. Third, explore the correspondence between inputs and outputs, for example, by modifying the inputs and observing their effects on the output. Fourth, calculate the gradient of the output to the inputs to know their contributions. It has also been suggested to combine the best of DL and fuzzy logic toward an enhanced interpretability \cite{Fan2021}.

\subsubsection{Bayesian approach to interpretability} 
\hypertarget{bayesian}
The work in \cite{Chakraborty} has discussed that there exist elements in DL and Bayesian reasoning that complement each other. Comparing Bayesian reasoning with DL, Bayesian reasoning offers a unified framework for modeling, inference, prediction, and decision making. Furthermore, uncertainty and variability of outcomes are explicitly accounted for. In addition, the framework has an "Occam's Razor" effect that penalizes overcomplicated models, which makes it robust to model overfitting. However, to ensure computational tractability, Bayesian reasoning is typically limited to conjugate and linear models.

In a recent survey on Bayesian DL (BDL) \cite{Wang2016} this complement observation has been exploited, and a general framework for BDL within a uniform probabilistic framework has been proposed. Further research is needed to be done to exploit this complement observation because it could improve model transparency and functionality \cite{Chakraborty}.

\subsubsection{Explaining competencies}
\hypertarget{competencies}
There is a need for users to gain a deeper understanding of the competencies of the AI system, which includes knowing what competencies it possesses, how its competencies can be measured, as well as whether or not it has blind spots (i.e., classes of solutions it never finds) \cite{gunning2019xai}. Through knowledge and competency research, XAI could play a significant role in society. Besides explaining to individuals, other roles include leveraging existing knowledge for further knowledge discovery and applications and teaching both agents and humans \cite{gunning2019xai}.

\subsection{Challenges and Research Directions of XAI in the Deployment Phase}
The following subsections are dedicated to challenges and research directions during the deployment of AI systems. The deployment phase starts from deploying ML solutions until we stop using the solutions (or maybe after that).

\subsubsection{Improving explanations with ontologies}
\hypertarget{ontologies}
An ontology is defined as \textit{"an explicit specification of a conceptualization"} \cite{GRUBER1993199}. The use of ontologies for representing knowledge of the relationships between data is helpful for understanding complex data structures \cite{Burkart2021-tq}. Therefore, the use of ontologies can help to produce better explanations as found in \cite{Panigutti,CONFALONIERI2021103471}.

The work in \cite{Burkart2021-tq} has discussed some recent works of the literature on this topic such as \cite{Panigutti,CONFALONIERI2021103471}. In \cite{Panigutti}, Doctor XAI was introduced as a model-agnostic explainer that focused on explaining the diagnosis prediction task of Doctor AI \cite{pmlr-v56-Choi16}, which is a black-box AI model that predicts the patient’s next visit time. It was shown that taking advantage of the temporal dimension in the data and incorporating the domain knowledge into the ontology helped improve the explanations' quality. Another work in \cite{CONFALONIERI2021103471} showed that ontologies can enhance human comprehension of global post-hoc explanations, expressed in decision trees.

It should be noted that ontologies are thought of as contributing a lot to explaining AI systems because they provide a user's conceptualization of the domain, which could be used as a basis for explanations or debugging \cite{tudorache2020ontology}. Toward that goal, new design patterns, new methodologies for creating ontologies that can support explainable systems, and new methods for defining the interplay between ontologies and AI techniques are needed \cite{tudorache2020ontology}. Furthermore, it is essential to conduct several user studies to determine the benefits of combining ontologies with explanations \cite{Burkart2021-tq}.

\subsubsection{XAI and privacy}
\hypertarget{privacy}
When individuals are affected by automated decision-making systems, two rights conflict: the right to privacy and the right to an explanation \cite{grant2020show}. At this stage, it could be a demand to disclose the raw training data and thus violate the privacy rights of the individuals from whom the raw training data came \cite{grant2020show}. Another legal challenge has been discussed in \cite{Longo2020}, which is the right to be forgotten \cite{VILLARONGA2018304}. By this right, individuals can claim to delete specific data so that they cannot be traced by a third party \cite{Longo2020}. Data preservation is another related issue because to use XAI to justify a decision reached by automated decision-making, the raw data used for training must be kept, at least until we stop using the AI solution.

One of the key challenges is establishing trust in the handling of personal data, particularly in cases where the algorithms used are challenging to understand \cite{Holzinger}. This can pose a significant risk for acceptance to end-users and experts alike \cite{Holzinger}. For example, end-users need to trust that their personal information is secured and protected as well as only their consented data is used, while experts need to trust that their input is not altered later \cite{Holzinger}.

Anonymization of data can be used to obscure the identity of people. However, privacy cannot always be protected by anonymization  \cite{grant2020show}. According to \cite{grant2020show}, the more information in a data set, the greater the risk of de-anonymization, even if the information is not immediately visible. Asserting that anonymization helps conceal who supplied the data to train the automated decision-making system might be comforting for the individuals whom the training data came from, but this does not the case with individuals who are entitled to an explanation of the results produced by the system \cite{grant2020show}.

In order to address some issues with anonymization techniques, it is recommended that further research should be undertaken in privacy-aware ML, which is the intersection between ML and security areas \cite{Holzinger}. XAI can play an essential role in this matter because to develop new techniques to ensure privacy and security, it will be essential to learn more about the inner workings of the system they are meant to protect \cite{Holzinger}. In addition, in the future, to promote the acceptance of AI and increase privacy protection, XAI needs to provide information on how the personal data of a particular individual was utilized in a data analysis workflow \cite{Longo2020}. However, according to \cite{grant2020show}, what if it is needed to review the data of many individuals and they may not have consented to review their data in litigation. In such cases, a path to review data for which individuals have not consented would be demanded, but it would be difficult to find such a path \cite{grant2020show}.

\subsubsection{XAI and security}
\hypertarget{security}
Two main concerns have been discussed for XAI and security: confidentiality and adversarial attacks \cite{arrieta2020,LIANG2021168,Ras2018,Tjoa2020}. For the confidentiality concern, several aspects of a model may possess the property of confidentiality \cite{arrieta2020}. As an example given by \cite{arrieta2020}, think of a company invested in a multi-year research project to develop an AI model. The model's synthesized knowledge may be regarded as confidential, and hence if only inputs and outputs are made available, one may compromise this knowledge \cite{orekondy19knockoff}. The work in \cite{joon18iclr} presented the first results on how to protect private content from automatic recognition models. Further research is recommended to develop XAI tools that explain ML models while maintaining models' confidentiality \cite{arrieta2020}.

Turning now to the adversarial attacks concern, the information revealed by XAI can be utilized in generating efficient adversarial attacks to cause security violations, confusing the model and cause it to produce a specific output, and manipulation of explanations \cite{arrieta2020,Tjoa2020}. In adversarial ML, three types of security violations can be caused by attackers using adversarial examples \cite{Huang2011}: integrity attacks (i.e., the system identifies intrusion points as normal), availability attacks (i.e., the system makes multiple classification errors, making it practically useless), and privacy violation (i.e., violating the privacy of system users). Attackers can do such security violations because an AI model can be built based on training data influenced by them, or they might send carefully crafted inputs to the model and see its results \cite{Huang2011}. According to \cite{LIANG2021168}, existing solutions to handle perturbations still suffer from some issues, including instabilities and lack of variability. Therefore, it is necessary to develop new methods to handle perturbations more robustly \cite{LIANG2021168,Ras2018}.

The information uncovered by XAI can also be utilized in developing techniques for protecting private data, e.g., utilizing generative models to explain data-driven decisions \cite{arrieta2020}. Two recent research directions have been highlighted in this context \cite{arrieta2020}: using generative models as an attribution method to show a direct relationship between a particular output and its input variables \cite{baumgartner2018visual}. The second is creating counterfactuals through generative models \cite{Liu2019}. It is expected that generative models will play an essential role in scenarios requiring understandable machine decisions \cite{arrieta2020}.

\subsubsection{XAI and safety}
\hypertarget{safety}
Trust and acceptance are benefits of explainability/interpretability \cite{NaisehJiang2020}. However, focusing on benefits without considering the potential risks may have severe consequences (e.g., relying too much or too little on the advice provided by the prescription recommendation system) \cite{NaisehJiang2020}. Several studies have been conducted to evaluate the safety of processes that depend on model outputs because erroneous outputs can lead to harmful consequences in some domains \cite{arrieta2020}. Therefore, possible risks must be at the top priority when designing the presented explanations \cite{NaisehJiang2020}.

Many techniques have been proposed to minimize the risk and uncertainty of adverse effects of decisions made using model outputs \cite{arrieta2020}. As an example, the model’s output confidence technique can examine the extent of uncertainty resulting from lack of knowledge regarding the inputs and the corresponding output confidence of the model to notify the user and cause them to reject the output produced by the model \cite{arrieta2020}. In order to achieve this, explaining what region of the inputs was used by the model to arrive at the outcome can be used for separating out such uncertainty that may exist within the input domain  \cite{arrieta2020}. Additionally, as has been suggested in \cite{NaisehJiang2020}, it is important to develop explanations that evolve with time, keeping in mind past explanations for long-term interactions with end-users and identifying ways to minimize risks. Developing evaluation metrics and questionnaires would be essential to integrate the user-centric aspects of explanations as well as evaluating error-proneness and any possible risks \cite{NaisehJiang2020}. Finally, in \cite{HUANG2020100270}, some major challenges have been discussed, including developing distance metrics that more closely reflect human perception, improvement to robustness by designing a set of measurable metrics for comparing the robustness of black-box AI models across various architectures, verification completeness using various verification techniques, scalable verification with tighter bounds, and unifying formulation of interpretability.

\subsubsection{Human-machine teaming}
\hypertarget{teaming}
Most provided explanations for AI systems are typically static and carry one message per explanation \cite{Abdul2018}. Explanations alone do not translate to understanding \cite{Adadi2018}. Therefore, for a better understanding of the system, users should be able to explore the system via interactive explanations, which is a promising research direction to advance the XAI field \cite{Abdul2018,Adadi2018}.

Even though there are already some works in this research direction as has been reported in \cite{Abdul2018}, much work is still needed to tailor interfaces to different audiences, exploit interactivity, and choose appropriate interactions for better visualization designs \cite{Abdul2018,Chatzimparmpas}. Various works have also been suggested to go beyond static explanations and enhance human-machine teaming. In \cite{Messina}, open-ended visual question answering (VQA) has been suggested to be used rather than providing a report with too many details. Here, an user queries (or make follow-up questions), and the system answers. Achieving that would provide better interaction between the system and the expert user. In another work \cite{Choo8402187}, it has been mentioned that generative models can allow for interactive DL steering because they allow for multiple answers. They highlighted that developing new DL models capable of adapting to various user inputs and generating outputs accordingly as well as developing visualization-based interfaces that enable effective interaction with DL systems are promising research areas in the future.

In \cite{Abdul2018}, rather than providing static explanations, the authors have suggested building on existing intelligibility work for context-aware systems (e.g., design space explorations, conceptual models for implicit interaction, and intelligible interfaces for various scenarios and using a variety of modalities). Additionally, they have highlighted a research area that is effectively interacting with AI augmentation tools. In \cite{Adadi2018}, it has been emphasized the importance of bridging HCI empirical studies with human sciences theories to make explainability models more human-centered models. In this way, adaptive explainable models would emerge by providing context-aware explanations that could be adapted to any changes in the parameters of their environment, such as user profile (e.g., expertise level, domain knowledge, cultural background, interests and preferences) and the explanation request setting (e.g., justification).

The authors in \cite{Chatzimparmpas} have mentioned that extracting, visualizing, and keeping track of the history of interaction data between users and systems can allow users to undo certain actions and examine them interactively would help to address some common challenges (e.g., hyperparameter exploration). Finally, the authors in \cite{NaisehJiang2020} have highlighted that user-friendliness and intelligent interface modalities need to take into account the type of explanations that meet users' goals and needs. For example, the system can ask for feedback from the users to know how good was the provided explanations (e.g., ``explain more'', ``redundant explanation'', or ``different explanation''). Such interaction can help to improve future explanations.

Taken together, it seems that different ways are needed to enhance human-machine teaming. Approaching HCI and other related studies can contribute to making explainability models more human-centered. In addition, humans can provide feedback on the provided explanations, which can help in improving future explanations.

\subsubsection{Explainable agency}
\hypertarget{agency}
Explainable agency refers to a general capability in which autonomous agents must provide explanations for their decisions and the reasons leading to these decisions \cite{Langley}. Based on the three explanation phases proposed in \cite{Neerincx}, the authors in \cite{Anjomshoae} presents a research roadmap for the explainable agency.

The first phase is explanation generation which is intended to explain why an action/result was taken/achieved \cite{Anjomshoae}. This phase of research focuses on the following key research directions: (1) there is a need to connect the internal AI mechanism of the agent/robot with the explanation generation module, (2) to produce dynamic explanations, new mechanisms are required for identifying relevant explanation elements, identifying its rationales, and combining these elements to form a coherent explanation.

The second phase is the explanation communication phase. Here, the focus is on what content end users will receive and how to present that content \cite{Neerincx}. According to \cite{Anjomshoae}, explainable agents/robots may be deployed in a variety of environments. Therefore, for some cases, multimodal explanation presentations (e.g., visual, audio, and expressive) could be a useful explanation communication approach for enabling efficient explainable agency communication.

For the last phase, explanation reception, the focus is on the human’s understanding of explanations. Some considerations should be taken into account to ensure an accurate reception \cite{Anjomshoae}. It is important to develop metrics to measure the explanations' effectiveness and the users' reaction to the provided explanations. In addition, the agent/robot should maintain a model of user knowledge and keep updating it based on the evolution of user expertise and the user's perception of the State of Mind (SoM) of the agent/robot, i.e., an internal representation of how the agent/robot treats the outer world.

\subsubsection{Machine-to-machine explanation}
\hypertarget{mexplanation}
A promising area of research is enabling machine-to-machine communication and understanding \cite{Weller2019}. Furthermore, it is an important research area because of the increasing adoption of the Internet of Things (IoT) in different industries. A growing body of research has begun exploring how multiple agents can efficiently cooperate and exploring the difference between explanations intended for humans and those intended for machines \cite{Preece2018,Weller2019}. 

According to \cite{Preece2018}, future explainable approaches are likely to provide both human and machine explanations, especially adaptive explainable approaches \cite{Adadi2018}. For machine explanations, complex structures that are beyond the comprehension of humans may be developed \cite{Weller2019}. However, how is it possible to measure the success of ``transfer of understanding'' between agents? The work in \cite{Weller2019} has suggested a metric for that, which is measuring the improvement of agent B’s performance on a particular task, or set of tasks, as a result of the information obtained from agent A - though it will be crucial to determine some key details, such as the bandwidth constraints and already existing knowledge with agent A.

Based on what has been mentioned above, it is expected that much work is going to be done on how to construct machine explanations, how to communicate these explanations, and which metrics we need to measure as a success of the transfer of understanding between agents and how to measure them. With more research into how machines communicate/explain themselves, we will be able to understand intelligence better and creating intelligent machines \cite{Molnar2021-do}.

\subsubsection{XAI and reinforcement learning}
\hypertarget{reinforcement}
The use of DL by reinforcement learning (RL) has been applied successfully to many areas \cite{dao2020demystifying}. Through the explicit modeling of the interaction between models and environments, RL can directly address some of the interpretability objectives \cite{Lipton2018-dx}. Despite that, unexplained or non-understandable behavior makes it difficult to users to trust RL agents in a real environment, especially when it comes to human safety or failure costs \cite{dao2020demystifying}. Additionally, we lack a clear understanding of why an RL agent decides to perform an action and what it learns during training \cite{dao2020demystifying}. RL's interpretability can help in exploring various approaches to solving problems \cite{dao2020demystifying}. For instance, understanding why the RL AlphaFold system \cite{senior2020improved} is capable of making accurate predictions can assist bioinformatics scientists in understanding and improving the existing techniques in protein structures to speed produce better treatment before new outbreaks happen \cite{dao2020demystifying}.

Recently, the work in \cite{Wells} highlighted several issues that need to be addressed and potential research directions in the area of XAI for RL. The authors find that the selected studies used "toy" examples or case studies that were intentionally limited in scope mainly to prevent the combinatorial explosion problem in the number of combinations of states and actions. Therefore, more focus on real-world applications has been suggested. It has also been mentioned that there is a lack of new algorithms in the area. Therefore, the design of RL algorithms with an emphasis on explainability is essential. Symbolic representations can be utilized so RL agents can inherently be explained and verified. Another issue is highlighted, which is the lack of user testing with the existing approaches, which is in line with what was mentioned in \cite{MILLER20191}. As for the complexity of the provided explanations, it has been found that the current focus is presenting explanations for users with a background in AI. Therefore, it has been suggested to conduct further research to present the explanations for those who might interact with the agents, which may have no background in AI. For example, providing more visceral explanations, e.g., annotations in a virtual environment. Additionally, enriching visualization techniques by considering the temporal dimensions of RL and multi-modal forms of visualization, e.g., virtual or augmented reality. Lastly, it has been emphasized the importance of open-source code sharing for the academic community.

Another interesting point for consideration has been highlighted in \cite{Guidotti2018-tx}, which is learning from explanations. The work in \cite{Krening} provides a starting point, which presents an agent who trained to simulates the Mario Bros. game using explanations instead of prior play logs.

\subsubsection{Explainable AI planning (XAIP)}
\hypertarget{planning}
Existing literature focuses mainly on explainability in ML, though similar challenges apply to other areas in AI as well \cite{Adadi2018}. AI planning is an example of such an area that is important in applications where learning is not an option \cite{fox2017explainable}. Recent years have seen increased interest in research on explainable AI planning (XAIP) \cite{Hoffmann2019}. XAIP includes a variety of topics from epistemic logic to ML, and techniques including domain analysis, plan generation, and goal recognition \cite{Hoffmann2019}. There are, however, some major trends that have emerged, such as plan explanations, contrastive explanations, human factors, and model reconciliation \cite{Hoffmann2019}.

Recently, the work in \cite{fox2017explainable} has explored the explainability opportunities that arise in AI planning. They have provided some of the questions requiring explanation. They also have described initial results and a roadmap toward achieving the goal of generating effective explanations. Additionally, they have suggested several future directions in both plan explanations and executions. Temporal planning, for instance, can open up interesting choices regarding the order of achieving (sub)goals. It is also interesting to consider whether giving the planner extra time to plan would improve the performance. In addition, one of the challenges in plan execution is explaining what has been observed at the execution time that prompts the planner to make a specific choice. As with XAI, it is crucial to have a good metric for XAIP that defines what constitutes a good explanation. Finally, it is imperative that the existing works on XAIP be reconsidered and leveraged so that XAIP will be more effective and efficient when used in critical domains.

\subsubsection{Explainable recommendation}
\hypertarget{recommendation}
Explainable recommendation aims to build models that produce high quality recommendations as well as provide intuitive explanations that can help to enhance the transparency, persuasiveness, effectiveness, trustworthiness, and satisfaction of recommendation systems \cite{ZhangINR-066}. 

The work in \cite{ZhangINR-066} conducted a comprehensive survey of explainable recommendations, and they discussed potential future directions to promote explainable recommendations. With regards to the methodology perspective, it has been suggested that (1) further research is needed to make deep models explainable for recommendations because we still do not fully understand what makes something recommended versus other options, (2) develop knowledge-enhanced explainable recommendation which allows the system to make recommendations based on domain knowledge, e.g., combine graph embedding learning with recommendation models, (3) use heterogeneous information for explainability such as multi-modal explanations, transfer learning over heterogeneous information sources, information retrieval and recommendation cross-domain explanations, and the impact that specific information modalities have on user receptiveness on the explanations, (4) develop context-aware explainable recommendations, (5) aggregate different explanations, (6) integrate symbolic reasoning and ML to make recommendations and explainability better by advancing collaborative filtering to collaborative reasoning, (7) further research is needed to help machines explain themselves using natural language, and (8) with the evolution of conversational recommendations powered by smart agent devices, users may ask “why” questions to get explanations when a recommendation does not make sense. Therefore, it is essential to answer the “why” in conversations which could help to improve system efficiency, transparency, and trustworthiness.

For the evaluation perspective, the authors in \cite{ZhangINR-066} have been suggested that the importance of developing reliable and easily implemented evaluation metrics for different evaluation perspectives (i.e., user perspective and algorithm perspective). Additionally, evaluating explainable recommendation systems using user behavior perspectives may be beneficial as well. Lastly, it has been highlighted that explanations should have broader effects than just persuasion. For example, investigate how explanations can make the system more trustworthy, efficient, diverse, satisfying, and scrutable.

In \cite{NaisehJiang2020}, the authors have presented several research challenges in delivery methods and modalities in user experience. As mentioned in that paper, for the delivery method, the current focus in the literature is on providing the explanation to the users while they are working on a task or looking for recommendations. However, more focus should be done on the long-term retrieval of such explanations, for example, through a digital archive, and their implications for accountability, traceability, and users' trust and adoption. That could increase the adoption of intelligent human-agent systems in critical domains. Another challenge is designing autonomous delivery capable of considering the context and situation in which users may need explanations and suitable explanations for them. It is worth mentioning that privacy matters should be taken into account when deriving the recommendations.

It has also been highlighted in \cite{NaisehJiang2020} that users' goals and needs would have to be met by user-friendly and intelligent interface modalities that provide appropriate explanations. Further, interaction with the system is needed and could help to improve future generated explanations. Finally, focusing on the benefits of explainability without considering the potential risks may have severe consequences. Therefore, when designing explanations, possible risks should be the first priority.

\subsubsection{XAI as a service}
\hypertarget{service}
There is an increasing trend in developing automated ML (AutoML) tools \cite{Molnar2021-do}. AutoML tool is an end-to-end pipeline starting with raw data and going all the way to a deployable ML model. Model-agnostic explanation methods are applicable to any ML model resulting from automated ML \cite{Molnar2021-do}. Similarly, we can automate the explanation step: calculate the importance of each feature, plot the partial dependence, construct a surrogate model, etc \cite{Molnar2021-do}. Some existing AutoML tools provide automatic generated explanations, e.g., AutoML H2O \cite{H2OAutoML20} and MLJAR AutoML \cite{mljar}. We expect that more Auto XAI tools will be available in the future, either incorporated with AutoML tools or as services.

\section{Conclusions}
In this systematic meta-survey paper, we present two main contributions to the literature of XAI. First, we propose an attempt to present a distinction between explainability and interpretability terms. Second, we shed light on the significant challenges and future research directions of XAI resulting from the selected 58 papers, which guide future exploration in the XAI area. Even though they are presented individually in 39 points, they can overlap and combine them based on researchers' backgrounds and interests, resulting in new research opportunities where XAI can play an important role. This meta-survey has three limitations. First, because we cannot ensure that the selected keywords are complete, we could miss some very recent papers. Second, to avoid listing the challenges and future research directions per each paper, we come up with the reported 39 points, which are the results of combining what was reported in the selected papers based on the authors' point of view. Third, we believe that more challenges and future research directions can be added where XAI can play an important role in some domains, such as IoT, 5G, and digital forensic. However, related surveys did not exist at the time of writing this meta-survey.

\bibliographystyle{unsrtnat}
\bibliography{references}  %%% Uncomment this line and comment out the ``thebibliography'' section below to use the external .bib file (using bibtex) .

\end{document}